\documentclass[runningheads]{llncs}
\usepackage[T1]{fontenc}
\usepackage{graphicx}
\usepackage{amsmath}
\usepackage{bm}
\usepackage{dsfont}
\usepackage{booktabs}
\begin{document}
\title{Evaluating and Calibrating Diffusion Model-derived Uncertainty for Quantitative MRI Mapping}
\titlerunning{Diffusion Model-derived Uncertainty for qMRI}
%
\author{Shishuai Wang \inst{1} \and 
Stefan Klein \inst{1} \and 
Juan A. Hernandez-Tamames \inst{1,2} \and 
Dirk H.J. Poot \inst{1}}
\authorrunning{S. Wang et al.}
%
\institute{Department of Radiology and Nuclear Medicine, Erasmus MC, Rotterdam, The Netherlands \and
Department of Imaging Physics, TU Delft, Delft, The Netherlands \\
\email{s.wang@erasmusmc.nl}}
\maketitle              
\begin{abstract}
Quantitative MRI (qMRI) provides standardised tissue parameter maps, but the reliability of deep learning-based qMRI mapping methods is often not explicitly characterised.
In this work we systematically evaluate uncertainty maps for quantitative MRI derived from multiple inferences of a data-consistent diffusion model-based qMRI framework.
Evaluation on synthetic test data assessed error-awareness, high-error detection, selective prediction, and Gaussian interval calibration.
Diffusion model-derived uncertainty was positively associated with the mapping error, while risk-coverage analysis showed that excluding high-uncertainty voxels reduced the retained error.
However, the raw uncertainty was poorly calibrated for quantitative interval interpretation.
Calibration was substantially improved using a post-hoc procedure combining prediction-value-dependent bias correction with scalar uncertainty scaling.
Qualitative evaluation on a healthy volunteer showed spatially meaningful uncertainty patterns.
These results indicate that diffusion model-derived uncertainty is informative for reliability assessment and selective prediction, but requires calibration for quantitative interval interpretation.
\end{abstract}

\section{Introduction}
Unlike conventional MRI, where image contrast is determined by a mixture of tissue properties and acquisition settings, quantitative MRI (qMRI) explicitly estimates biophysical tissue parameters such as proton density and relaxation times. 
The resulting quantitative maps offer a more standardised representation of tissue properties, which is desirable for cross-subject analysis, multi-centre studies, and longitudinal monitoring.
Quantitative MRI typically requires the acquisition of multiple contrast-weighted images with varying sequence parameters, from which quantitative maps are subsequently estimated. 
Conventional estimation approaches rely on signal models, such as non-linear model fitting or dictionary matching, while recent deep learning-based methods have shown promising performance. 
However, uncertainty estimation, which is important for assessing the reliability of the resulting quantitative maps, is often not routinely provided or systematically evaluated.

For deep learning-based methods, epistemic uncertainty can be estimated using approaches such as Monte Carlo dropout~\cite{gal2016dropout} or model ensembles~\cite{lakshminarayanan2017simple,ovadia2019can,zhao2022efficient,wang2023uncertainty}. 
Nevertheless, these methods typically require architectural modifications, additional model training, or specialized inference strategies. 
In contrast, diffusion models have recently shown promise for qMRI mapping~\cite{wang2024qmri,bian2024diffusion,wang2026q3}. 
Due to their stochastic sampling process, diffusion models can produce different predictions from the same input, making the standard deviation from repeated inference results a straightforward way to estimate uncertainty without additional model modifications or retraining~\cite{chung2022score}.
Previous work has preliminarily shown that the diffusion model-derived uncertainty visually overlaps with error-prone regions~\cite{wang2024qmri}. 
However, it remains unclear whether such uncertainty estimates are quantitatively associated with mapping errors, whether they can support selective prediction, and whether they are calibrated when interpreted as Gaussian prediction intervals.

In this work, we systematically evaluate diffusion model-derived uncertainty for quantitative T1 and T2 mapping using a data-consistent diffusion-based qMRI framework.
Mean and standard deviation of repeated inferences are used as output quantitative map and uncertainty map respectively.
The evaluation is performed on synthetic data with available ground truth, and an additional real healthy volunteer scan used for qualitative visualisation. 
To summarise, our contributions are:
\begin{itemize}
\item We provide a systematic evaluation of diffusion model-derived uncertainty in qMRI, covering error-awareness, high-error detection, selective prediction, and coverage calibration.
\item We show that sampling uncertainty is informative for reliability assessment and uncertainty-guided selective prediction, but is not inherently calibrated as a Gaussian prediction interval.
\item We investigate a simple post-hoc calibration strategy based on prediction-value-dependent bias correction and scalar uncertainty scaling, which improves Gaussian interval calibration on the test dataset.
\end{itemize}

\section{Materials and Methods}
\subsection{Diffusion model-derived uncertainty}
A diffusion model-based qMRI mapping framework \cite{wang2026q3} is used as a testbed in this study. 
The model is based on a denoising diffusion probabilistic model integrated with data-consistency enforcement during inference.
The input to the model is the stacked multi-echo weighted images.
For each input, the stochastic sampling process was repeated $K$ times, yielding predictions $\hat{\bm{y}}^{(1)},\ldots,\hat{\bm{y}}^{(K)}$. 
For each parameter $p \in \{T_1, T_2\}$, the mean prediction and diffusion model-derived uncertainty are defined as
\begin{equation}
    \mu_p=\frac{1}{K}\sum_{k=1}^{K}\hat{y}_p^{(k)}, \; \text{and}\; \sigma_p=\sqrt{\frac{1}{K-1}\sum_{k=1}^{K}\left(\hat{y}^{(k)}_p-\mu_p\right)^2}
\end{equation}
respectively.
Given the ground-truth map $y_p$, the absolute error map and standardised residual are defined as
\begin{equation}
    e_p := |y_p - \mu_p|, \; \text{and} \; r_p:=\frac{y_p - \mu_p}{\sigma_p}
\end{equation}
respectively. 
Voxels with $\sigma_p \leq 10^{-8}$ were excluded from the standardised residual and calibration correction for numerical stability.
All operations are applied voxel-wise.

\subsection{Experimental data}
Training, calibration and test datasets were generated from 20 BrainWeb digital brain phantoms~\cite{cocosco1997brainweb}. 
Nineteen phantoms were used for model training and post-hoc calibration, while the remaining phantom was held out for testing.
For model training, five independently generated quantitative-map realisations were created for each of the 19 training phantoms.
For post-hoc calibration, one additional realisation was generated for each of these 19 phantoms and these realisations were not used for model training.
For testing, 20 independently generated realisations were created from the held-out phantom.
For each realisation of quantitative maps, noisy k-space data were generated and the corresponding weighted images were reconstructed.
The model training and inference hyperparameters were kept the same as \cite{wang2026q3}.
During inference, the selected qMRI framework was applied $K=10$ times to compute $\mu_p$ and $\sigma_p$.

In addition, one previously acquired healthy volunteer scan using the same qMRI protocol was included for qualitative visualisation.
The scan was acquired on a 3.0T GE SIGNA Premier scanner using a 48-channel head coil.
Since voxel-wise ground-truth quantitative maps are unavailable in vivo, this scan was not included in the quantitative evaluation.

\subsection{Evaluation metrics}
All metrics were computed within the foreground head mask and reported for each parameter as mean $\pm$ standard deviation across the test dataset.
Mapping accuracy was evaluated using the mean absolute error (MAE) and root mean squared error (RMSE) between $\mu_p$ and $y_p$.
The magnitude of diffusion model-derived uncertainty was summarised using the mean and 95th percentile of $\sigma_p$.

To assess error-awareness, we computed the Spearman correlation between $e_p$ and $\sigma_p$.
As an indication of high-error detection performance, we computed the area under the receiver operating characteristic curve (AUROC) when using $\sigma_p$ to detect the 10\% voxels with the highest $e_p$.

Selective prediction was evaluated using risk-coverage analysis.
Voxels were sorted from low to high uncertainty.
For a retained coverage fraction $c$, only the $c$ fraction of voxels with the lowest uncertainty was retained, and the retained risk was computed as the MAE over these voxels.
The area under the risk-coverage curve (AURC) was used as a summary measure, where lower values indicate better selective prediction.
For comparison, we also computed a random-ranking baseline, representing chance-level voxel selection, and an oracle-error-ranking baseline, where voxels were sorted from low to high true absolute error.

Calibration was evaluated by interpreting the uncertainty estimate as a Gaussian predictive standard deviation.
For a nominal central coverage level $q\in(0,1)$, the corresponding Gaussian interval was defined using
\begin{equation}
    z_q=\Phi^{-1}\left(\frac{1+q}{2}\right),
\end{equation}
where $\Phi$ denotes the standard normal cumulative distribution function.
The empirical coverage probability was computed as
\begin{equation}
    CP(q) = \frac{1}{|\Omega|}\sum_{x\in\Omega}
    \mathbf{1}(|y_p(x)-\mu_p(x)| \leq z_q\sigma_p(x)).
\end{equation}
where $\Omega$ denotes the foreground mask and $\mathbf{1}(\cdot)$ is the indicator function.
The coverage probability accuracy at nominal coverage level $q$ was defined as
\begin{equation}
    CPA(q)=|CP(q)-q|,
\end{equation}
with lower values indicating better interval calibration.
We used the central 50\% interval as the main scalar calibration metric, denoted as $CPA_{50}$ for brevity, corresponding to $q=0.5$ and $z_{0.5}=\Phi^{-1}(0.75)=0.674$.
Calibration curves were additionally evaluated over multiple nominal coverage levels.

\subsection{Post-hoc calibration}
To investigate whether miscalibration could be mitigated, we applied a simple post-hoc calibration procedure using the calibration dataset.
For each parameter $p$, a prediction-value-dependent bias correction was first estimated.
Calibration voxels were binned according to their predicted value $\mu_p$ using 50 quantile bins, and the median residual $y_p-\mu_p$ was computed within each bin.
This yielded an empirical lookup table $\hat{b}_p(\mu_p)$, which was linearly interpolated and applied to correct the mean prediction:
\begin{equation}
\mu_{p,\mathrm{corr}} = \mu_p+\hat{b}_p(\mu_p).
\end{equation}
Predictions outside the calibrated range were assigned the nearest boundary correction value.

After bias correction, a scalar uncertainty scaling factor was estimated to improve central interval calibration.
For the nominal central 50\% Gaussian interval, the scaling factor was estimated as
\begin{equation}
    \lambda_p = \frac{\mathrm{median}\left(|y_p-\mu_{p,\mathrm{corr}}|/\sigma_p\right)}{z_{0.5}},
\end{equation}
where the median was computed over foreground calibration voxels.
The calibrated uncertainty was then defined as
\begin{equation}
\sigma_{p,\mathrm{corr}}=\lambda_p\sigma_p.
\end{equation}

The lookup table $\hat{b}_p(\mu_p)$ and scaling factor $\lambda_p$ were estimated on the calibration dataset and then fixed when applied to the held-out test dataset.
On the test dataset, calibration was evaluated for three variants: the raw prediction, the bias-corrected prediction, and the bias-corrected prediction with scaled uncertainty.
The same fixed $\hat{b}_p(\mu_p)$ and $\lambda_p$ were used for all nominal coverage levels, and no additional calibration was performed for individual values of $q$.

\section{Results}
\begin{figure}[htb]
    \centering
    \includegraphics[width=\linewidth]{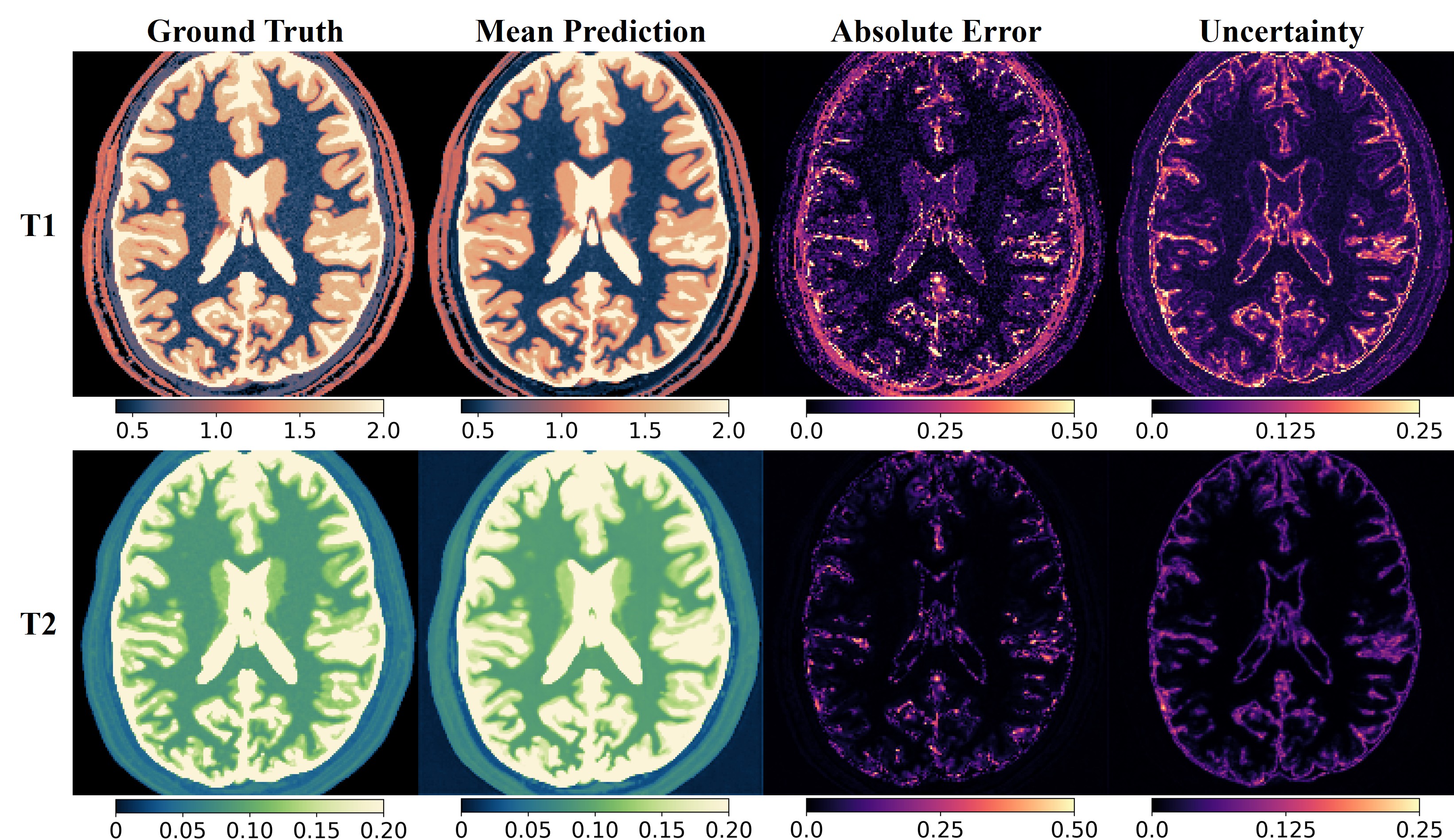}
    \caption{Representative qualitative results on synthetic test data. The unit is seconds.}
    \label{fig1}
\end{figure}

\begin{figure}[htb]
    \centering
    \includegraphics[width=0.85\linewidth]{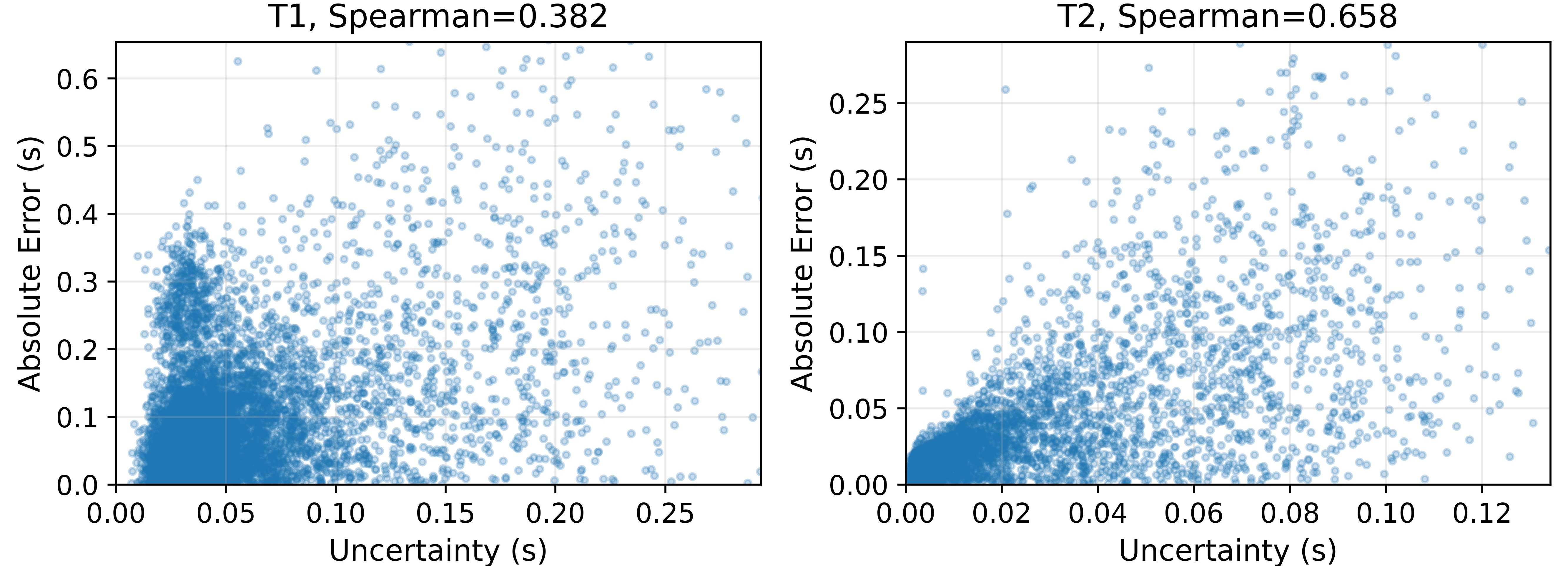}
    \caption{Voxel-wise association between absolute mapping error and diffusion model-derived uncertainty for one representative case. Spearman correlations were computed using all valid foreground voxels, while points were randomly subsampled for visualisation.}
    \label{fig2}
\end{figure}

\begin{figure}[htb]
    \centering
    \includegraphics[width=0.85\linewidth]{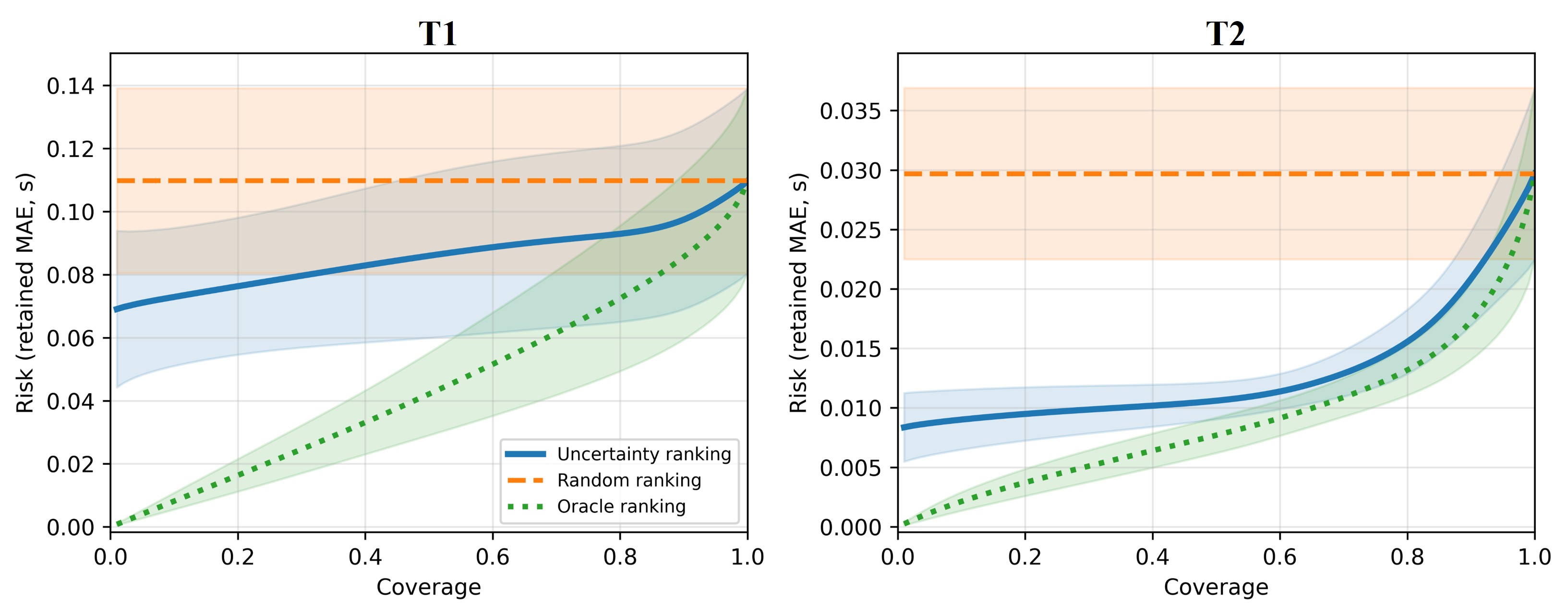}
    \caption{Risk-coverage analysis for uncertainty-guided selective prediction. Voxels were retained from low to high diffusion model-derived uncertainty. Random ranking represents chance-level selection, while oracle ranking retains voxels from low to high true absolute error and therefore provides an unattainable lower bound. Solid lines and shaded regions indicate mean $\pm$ standard deviation across 20 synthetic test realisations.}
    \label{fig3}
\end{figure}

A representative synthetic data example is shown in Fig.~\ref{fig1}. 
Elevated uncertainty was predominantly observed around tissue interfaces and in regions exhibiting increased absolute mapping error.
Table~\ref{tab1} summarizes the quantitative evaluation results. 
Diffusion model-derived uncertainty was positively associated with mapping errors. 
The uncertainty-error Spearman correlation was $0.260\pm0.095$ for T1 and $0.612\pm0.113$ for T2.
The representative scatter plots in Fig.~\ref{fig2} further illustrate this association.
Uncertainty also detected the highest-error voxels, yielding AUROC values of $0.773\pm0.076$ and $0.953\pm0.010$, respectively. 
As shown in Fig.~\ref{fig3}, excluding high-uncertainty voxels reduced the retained MAE, showing that diffusion model-derived uncertainty can support uncertainty-guided selective prediction.
Compared with the baselines, the uncertainty-guided curves were consistently below random ranking and above oracle-error ranking for both T1 and T2. 
This indicates that the uncertainty estimates provide useful, although non-optimal, voxel-wise reliability ranking.

\begin{table}[htb]
\centering
\caption{Summary of Uncertainty Evaluation on Synthetic Test Data}
\label{tab1}
\begin{tabular}{@{}ccccc@{}}
\toprule
Parameter & MAE $\downarrow$        & RMSE $\downarrow$ & Mean Uncertainty     & p95 Uncertainty            \\ \midrule
T1 (s)        & 0.110$\pm$0.029        & 0.148$\pm$0.034  & 0.059$\pm$0.003      & 0.166$\pm$0.010            \\
T2 (s)        & 0.030$\pm$0.007        & 0.054$\pm$0.017  & 0.017$\pm$0.002      & 0.077$\pm$0.008            \\ \midrule
Parameter & U-E Spearman $\uparrow$ & AUROC $\uparrow$  & AURC $\downarrow$ & Raw $CPA_{50}$ $\downarrow$ \\ \midrule
T1        & 0.260$\pm$0.095        & 0.773$\pm$0.076  & 0.079$\pm$0.022      & 0.287$\pm$0.064            \\
T2        & 0.612$\pm$0.113        & 0.953$\pm$0.010  & 0.012$\pm$0.001      & 0.370$\pm$0.033            \\ \bottomrule
\end{tabular}
\end{table}

\begin{figure}[htb]
    \centering
    \includegraphics[width=0.85\linewidth]{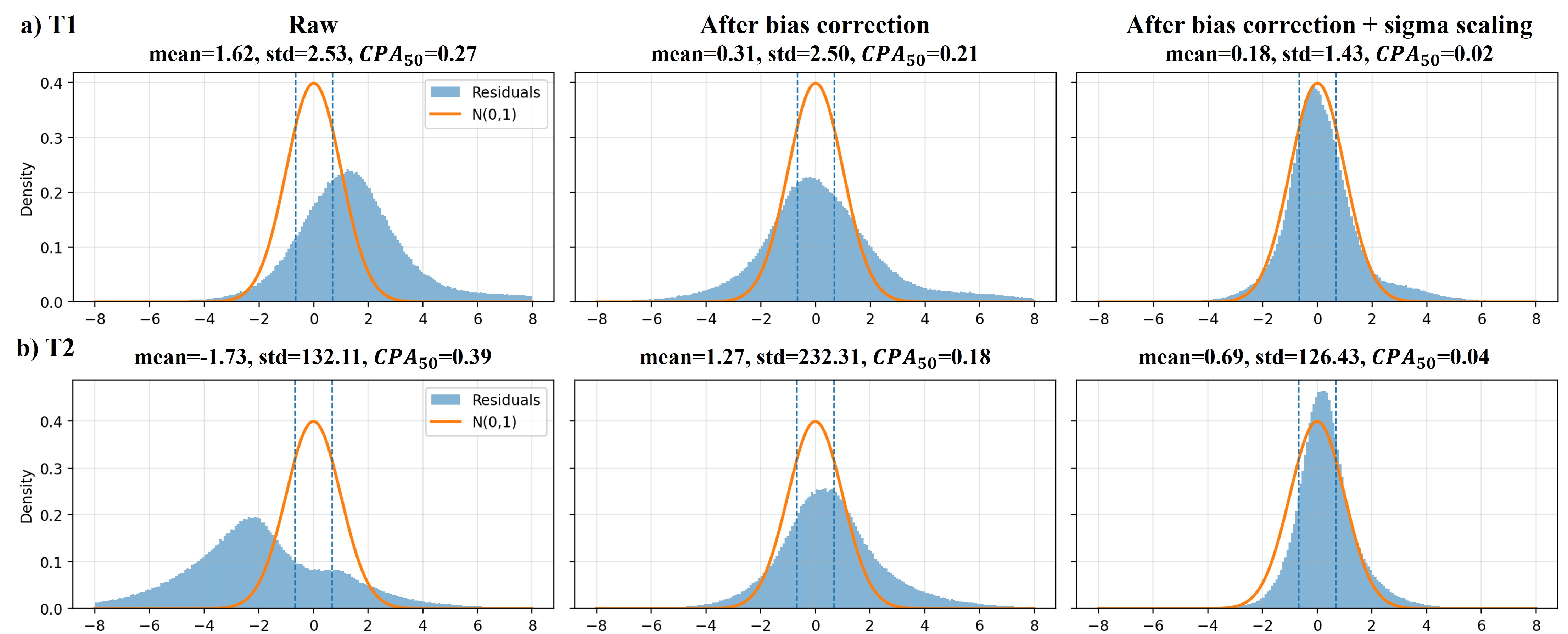}
    \caption{Standardised residual distributions before and after post-hoc calibration for one representative case. The standard normal density is shown for reference, and the dashed vertical lines indicate the central 50\% Gaussian interval. }
    \label{fig4}
\end{figure}

Despite its error-awareness and selective utility, the raw uncertainty showed poor interval calibration. 
The raw $CPA_{50}$ was $0.287\pm0.064$ for T1 and $0.370\pm0.033$ for T2, indicating substantial deviations from the nominal 50\% coverage. 
As summarized in Table~\ref{tab2}, the prediction-value-dependent bias correction reduced the MAE from $0.110\pm0.029$~s to $0.089\pm0.018$~s for T1, and from $0.030\pm0.007$~s to $0.023\pm0.006$~s for T2. 
It also reduced $CPA_{50}$ to $0.228\pm0.044$ and $0.240\pm0.039$, respectively. 
Subsequent uncertainty scaling using $\lambda_p$ further reduced $CPA_{50}$ to $0.063\pm0.054$ for T1 and $0.055\pm0.042$ for T2.
Figure~\ref{fig4} visualizes the standardised residual ($r_p$) distributions for one representative case. 
The raw residuals were shifted and substantially broader than the standard normal distribution. 
Bias correction reduced the central offset, while uncertainty scaling further improved the nominal 50\% interval coverage. 
Nevertheless, heavy-tailed residuals remained, particularly for T2, indicating that the post-hoc procedure improved central coverage calibration without making the complete residual distribution Gaussian.

\begin{table}[htb]
\centering
\caption{Post-hoc Calibration Correction Summary}
\label{tab2}
\begin{tabular}{ccccccc}
\hline
Parameter & $\lambda_p$ & \begin{tabular}[c]{@{}c@{}}Raw \\ MAE $\downarrow$\end{tabular} & \begin{tabular}[c]{@{}c@{}}Bias-corr.\\ MAE $\downarrow$\end{tabular} & \begin{tabular}[c]{@{}c@{}}Raw \\ $CPA_{50}$ $\downarrow$\end{tabular} & \begin{tabular}[c]{@{}c@{}}Bias-corr.\\ $CPA_{50}$ $\downarrow$\end{tabular} & \begin{tabular}[c]{@{}c@{}}Bias+scale-corr.\\ $CPA_{50}$ $\downarrow$\end{tabular} \\ \hline
T1        & 1.750        & 0.110$\pm$0.029                                                 & 0.089$\pm$0.018                                                       & 0.287$\pm$0.064                                                        & 0.228$\pm$0.044                                                              & 0.063$\pm$0.054                                                                    \\
T2        & 1.837        & 0.030$\pm$0.007                                                 & 0.023$\pm$0.006                                                       & 0.370$\pm$0.033                                                        & 0.240$\pm$0.039                                                              & 0.055$\pm$0.042                                                                    \\ \hline
\end{tabular}
\end{table}

\begin{figure}[htb]
    \centering
    \includegraphics[width=0.85\linewidth]{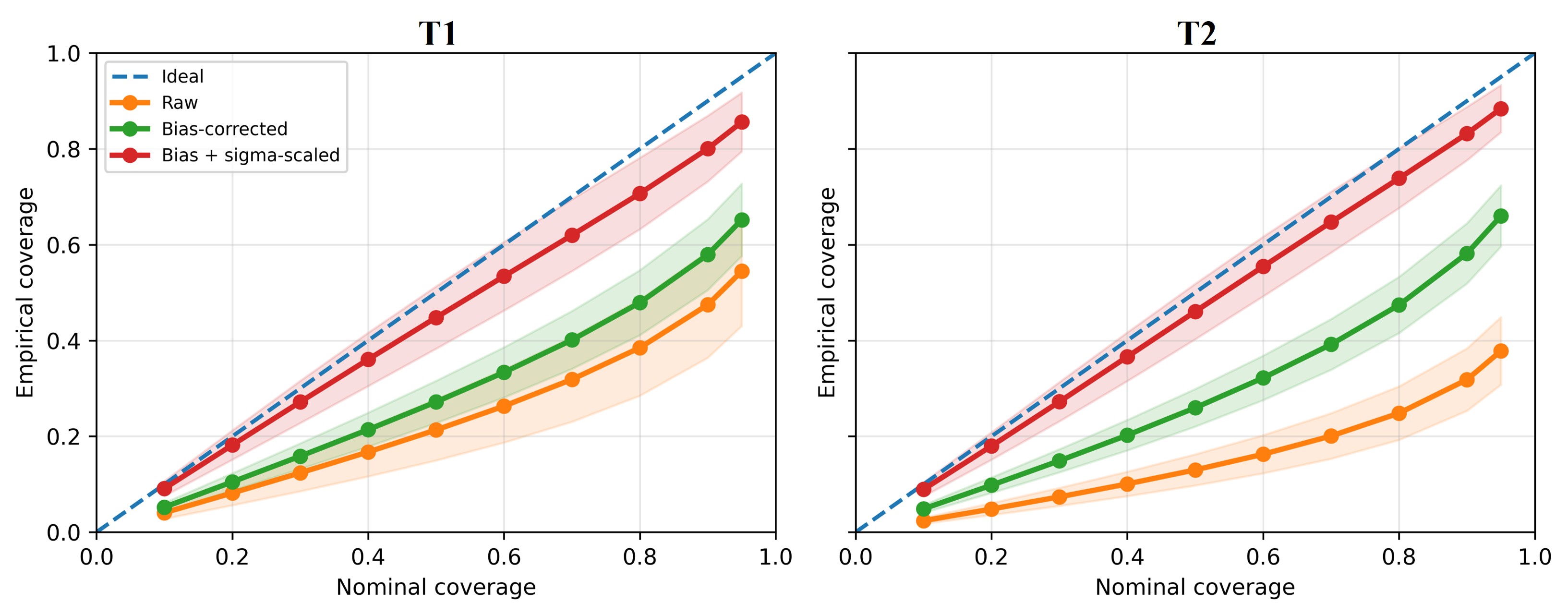}
    \caption{Calibration curves over multiple nominal coverage levels. The lookup-table bias correction and scalar uncertainty scaling were estimated on the calibration set and then fixed when applied to the test data. The scalar scaling factor was estimated to match the central 50\% interval. Lines and shaded regions indicate mean $\pm$ standard deviation across 20 test realisations.}
    \label{fig5}
\end{figure}

The calibration curves in Fig.\ref{fig5} further confirmed this behaviour across multiple nominal coverage levels. 
The raw intervals under-covered across the evaluated range, bias correction improved empirical coverage, and subsequent scalar uncertainty scaling shifted the curves closer to the ideal diagonal, particularly around the central coverage range targeted by the scaling procedure. 
Residual under-coverage at higher nominal coverage levels indicates that the post-hoc correction improved central interval calibration but did not fully calibrate the predictive distribution.

\begin{figure}[htb]
    \centering
    \includegraphics[width=0.75\linewidth]{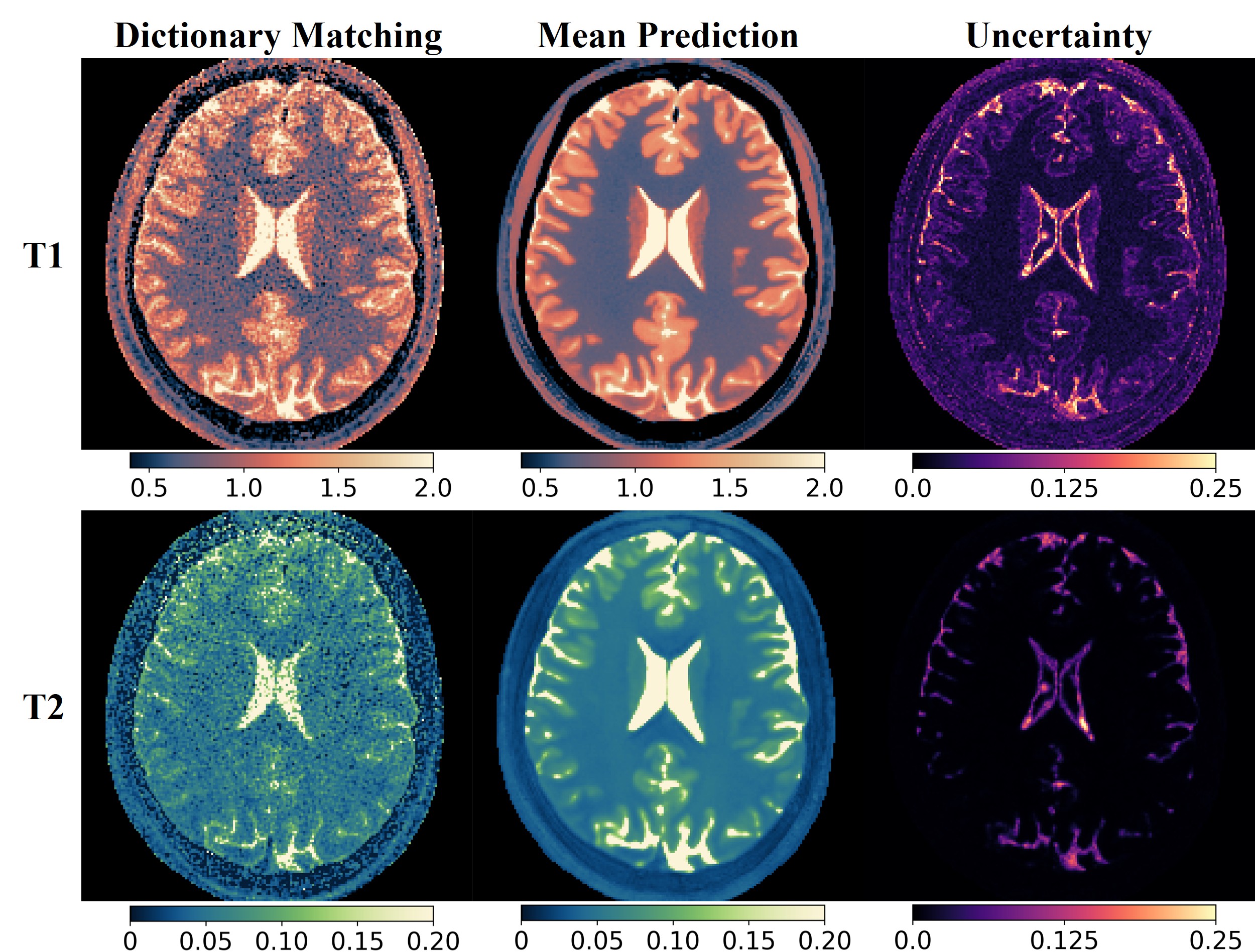}
    \caption{Qualitative results on a healthy volunteer scan. The unit is seconds.}
    \label{fig6}
\end{figure}

Finally, Fig.~\ref{fig6} shows qualitative results on in vivo data. 
Dictionary matching was included as a conventional model-based reference for qualitative comparison, but was not treated as ground truth.
The uncertainty maps showed elevated values mainly around tissue interfaces, and within CSF and grey matter. 
The real-data example was used for qualitative illustration only, as ground truth was unavailable.

\section{Discussion}
In this work, we treated the standard deviation across repeated diffusion model samples as diffusion model-derived uncertainty and systematically evaluated it for quantitative MRI mapping.
The results show that this uncertainty contains meaningful information about prediction reliability: it is positively associated with mapping errors, detected high-error regions, and supported uncertainty-guided selective prediction. 
However, it was not inherently calibrated when interpreted as a Gaussian predictive standard deviation. 
A simple post-hoc procedure combining prediction-value-dependent bias correction and scalar uncertainty scaling substantially improved the calibration on test data.

The different evaluation metrics capture complementary aspects of uncertainty quality. 
AUROC evaluates whether uncertainty ranks voxels in the tail of the error distribution above lower-error voxels. 
The strong high-error detection performance, especially for T2, could be used to flag or reject highly unreliable estimation in practice.
The positive uncertainty-error correlations further indicate a continuous association between uncertainty and error magnitude, while the risk-coverage results demonstrate that this ranking can be used to reduce retained error by excluding uncertain voxels.

Despite this error-awareness, the raw uncertainty substantially under-covered the nominal 50\% Gaussian prediction intervals. 
This indicates that such uncertainty estimation may be useful for ranking predictions without having the correct absolute scale for interval estimation. 
The standardised residual distributions were shifted and broader than a standard normal distribution, indicating contributions from both systematic prediction bias and underestimated uncertainty magnitude. 
The prediction-value-dependent lookup table reduced systematic bias and improved the accuracy of the mean quantitative maps. 
Subsequent scalar uncertainty scaling corrected the remaining scale mismatch and markedly reduced $CPA_{50}$ on the test data.
However, the post-hoc correction should not be interpreted as making the full predictive distribution Gaussian. 
Heavy-tailed standardised residuals remained, particularly for T2. 
The proposed scaling was derived specifically to match the nominal 50\% interval and therefore calibrates the central part of the residual distribution rather than its extremes.

Several limitations should be considered. 
Quantitative evaluation was limited to synthetic data, as ground truth is unavailable for in-vivo qMRI.
Although the calibration dataset used newly generated quantitative-map realisations, the underlying anatomies had been seen by the model during training. 
The uncertainty estimate was computed from $K=10$ repeated diffusion model inferences, which was chosen as a practical compromise between computational cost and repeated-sampling stability. 
The sensitivity of the reported uncertainty metrics to $K$ was not systematically investigated.
Moreover, this sampling-based uncertainty may not capture all sources of predictive uncertainty, including model bias and acquisition-related domain shift. 
The healthy-volunteer experiment therefore provides qualitative evidence only. 
Validation on additional anatomies, acquisition protocols, and pathological cases is required in future work.

Overall, diffusion model-derived sampling uncertainty was informative for identifying unreliable qMRI estimations and enabling selective prediction, but its raw magnitude should not be interpreted directly as a calibrated Gaussian predictive standard deviation. 
The results suggest that simple post-hoc bias and scale correction can substantially improve central interval calibration while preserving the practical error-awareness of the uncertainty estimate.

\begin{credits}
\subsubsection{\ackname} This work was conducted within the “Trustworthy AI for MRI” ICAI lab within the project ROBUST, funded by the Dutch Research Council (NWO), GE Healthcare, and the Dutch Ministry of Economic Affairs and Climate Policy (EZK).

\subsubsection{\discintname}
Shishuai Wang, Stefan Klein, Juan-Antonio Hernandez-Tamames and Dirk Poot received research grants from GE Healthcare. 
\end{credits}

%
\bibliographystyle{splncs04}
\bibliography{references}

\end{document}